# Deep Neural Network-Driven Adaptive Filtering

Qizhen Wang, Gang Wang and Ying-Chang Liang, *Fellow, IEEE*

*Abstract*—This paper proposes a deep neural network (DNN)-driven framework to address the longstanding generalization challenge in adaptive filtering (AF). In contrast to traditional AF frameworks that emphasize explicit cost function design, the proposed framework shifts the paradigm toward direct gradient acquisition. The DNN, functioning as a universal nonlinear operator, is structurally embedded into the core architecture of the AF system, establishing a direct mapping between filtering residuals and learning gradients. The maximum likelihood is adopted as the implicit cost function, rendering the derived algorithm inherently data-driven and thus endowed with exemplary generalization capability, which is validated by extensive numerical experiments across a spectrum of non-Gaussian scenarios. Corresponding mean value and mean square stability analyses are also conducted in detail.

*Index Terms*—Deep Neural Network (DNN), Adaptive Filtering, Data-Driven, Generalization Capacity.

## I. INTRODUCTION

**T**HE adaptive filtering (AF) technique has become ubiquitous in modern signal processing [1]–[5]. Current AF methods are fundamentally governed by additive noise statistics. In Gaussian environments, mean square error (MSE)-based solutions like the least mean square (LMS) yield the best results; In non-Gaussian environments, several AF frameworks have been proposed. Typical examples are Information Theoretical Learning (ITL) framework for super-Gaussian noise [6], [7], higher-order moment estimation-based framework for sub-Gaussian noise [8], and Gaussian Mixture Model (GMM)-based framework for multi-peak noise [9], [10].

Note that the above prevailing AF frameworks generally incorporate two critical design steps:

**Step 1:** Formulate an explicit cost function with a well-defined mathematical representation;

**Step 2:** Derive the gradient-based algorithm by differentiating the cost function.

This design paradigm is imbued with a tacit requirement: the cost function formulated in **Step 1** must be analytically tractable, otherwise it would impede the differentiation process in **Step 2**. This inherent constraint rigorously regulates the AF algorithm, rendering it mathematically determined by the structure of the cost function and consequently model-based. While model-based algorithms are optimal for specific noise profiles, they are intrinsically plagued by limited generalizability, suffering substantial performance degradation when confronted with distributional mismatches. Representative cases are the maximum correntropy criterion (MCC) [11], [12] and minimum error entropy (MEE) [13]–[15], which excel in

Q. Wang is with the National Key Laboratory of Wireless Communications, University of Electronic Science and Technology of China (UESTC), Chengdu 611731, China (e-mail: 202421220310@std.uestc.edu.cn).

G. Wang is with the School of Information and Communication Engineering, University of Electronic Science and Technology of China, Chengdu 611731, P.R. China (e-mail: wanggang_hld@uestc.edu.cn).

Y.-C. Liang is with the Center for Intelligent Networking and Communications (CINC), University of Electronic Science and Technology of China (UESTC), Chengdu 611731, China (e-mail: liangyc@ieee.org).

heavy-tailed noise but falter in light-tailed noise [16], [17]. Notwithstanding extensive efforts to improve generalization capabilities, developing AF algorithms adaptable to diverse noise characteristics remains an unresolved problem in practice. In this paper, we delve into this longstanding challenge.

To tackle this challenge, recent advances have introduced a radial basis function neural network (RBFNN)-based AF framework [18], where filtering residuals serve as network inputs and cost function values from the ITL act as outputs, demonstrating adaptability in varied non-Gaussian settings. Building on this, [19] employs RBFNNs to explicitly characterize the probability density function (PDF) of the noise, thereby deriving a novel algorithm, called RBF-AF, with superior generalization capabilities across arbitrary distributions.

While effective in theory, when deployed in real-world AF systems, the RBFNN exhibits substantial limitations in characterizing intricate distributions, for example, pronounced sensitivity to initial hyperparameter configuration [20] and difficulties in optimizing the cardinality of RBF units [21]. As such, its explicit formulation frequently suffers from accuracy deterioration, with non-negligible instances of prominent divergence that severely compromise algorithmic stability. This observation motivates the integration of deeper neural architectures to achieve more robust representation of nonlinear relationships in AF systems, thereby ensuring consistent generalization performance across diverse operational regimes.

Nevertheless, the "black-box" characteristics of deep neural networks (DNNs) stand in direct opposition to the canonical AF paradigm mandating explicit cost function formulation outlined in **Step 1**. At its core, the iterative optimization process in AF demands mathematically tractable cost functions for gradient computation. The representational nature of DNNs precludes this formulation, consequently obstructing the differentiation procedure in **Step 2**. Such inherent design constraints impose formidable obstacles to effective DNN incorporation.

In this paper, we propose a novel framework to address this design flaw. Leveraging the DNN, the nonlinear relationship between filtering residuals and learning gradients is directly bridged by a data-driven (DD) approach, thereby circumventing the need for explicit mathematical formulation of the cost function. Then the RBFNN cease to be the sole alternative. Deeper neural architectures can be integrated into the AF system to ensure enhanced superior generalization capability.

Specifically, the maximum likelihood serves as the implicit cost function, whose derivative produces the gradient for adaptive learning, which in turn act as the target for the DNN's output. Then two datasets are required for DNN training:

- **Dataset 1**: Measured samples of noise from the system.
- **Dataset 2**: A corresponding set of PDF derivative values.

Regarding **Dataset 1**, multiple approaches are available, including direct measurement methods [9], [10], [19] and indirect estimation techniques [22], [23]. **Dataset 2** is distilled from **Dataset 1** through machine learning techniques. A prevalent solution is kernel density estimation (KDE), a widely-adopted nonparametric method for probability distribution fitting [24]. Following the establishment of both datasets, a DNN is employed to characterize their intrinsic relationship. Thus, the learning gradient can be directly acquired through the DNN's feedforward computation, thereby completely circumventing analytical approaches. This constitutes a new AF paradigm.

To better position the contributions of this paper, we provide an organized classification of existing neural network (NN)-driven AF methods in Table I. Our innovation lies in **item (iii)**: the direct end-to-end learning of gradient operators instead of cost functions, which is a new unexplored dimension in AF.

TABLE I: Classification of existing NN-driven AF methods

| NN Input | NN Output | Output type |
|---|---|---|
| Residual | (i) ITL Value [18] | Cost Function Value |
| | (ii) PDF Value [19] | Cost Function Value |
| | (iii) Derivative value of the PDF | Gradient Value |

Capitalizing on the robust grounded capacity of the DNN as universal nonlinear approximators, the proposed algorithm achieves consistent performance advantages over advanced alternatives in intricate scenes encompassing impulse, uniform, skewed, and multi-peak, as demonstrated by numerical results. Furthermore, comprehensive analyses of the mean value and mean square stability properties are also conducted in detail.

The main contributions of this letter are twofold:

i) DNNs are structurally embedded into the core architecture of the AF system, giving rise to a novel algorithm that exhibits superior generalization capability compared to advanced methods. To the best of our knowledge, this pioneers the first native integration of DNNs into the AF field.

ii) A innovative AF framework is presented. Parallel to the classical AF framework's focus on the design of the cost function, the proposed AF framework specializes in direct gradient acquisition, establishing a genuinely DD architecture.

## II. PROPOSED DNN-AF ALGORITHM

Considering a linear filtering problem, where the goal is to estimate an unknown parameter $\boldsymbol{w}_o \in \mathbb{R}^M$. The system observes scalar measurements $d_i \in \mathbb{R}^1$ at each instant $i$, generated through the following regression model

$$d_i = \boldsymbol{w}_o^T \boldsymbol{u}_i + v_i, \forall i \tag{1}$$

where the system input $\boldsymbol{u}_i$ is white noise with zero mean and variance $\sigma_u^2$, and the additive noise $v_i$ has a variance of $\sigma_v^2$.

The filter residual $e_i$ at instant $i$ is obtained by

$$e_i = d_i - \boldsymbol{w}_i^T \boldsymbol{u}_i, \tag{2}$$

where $\boldsymbol{w}_i$ is the estimate of parameter $\boldsymbol{w}_o$.

When the mean square error (MSE) criterion is employed, we arrive at the renowned LMS algorithm, as shown in (3)

$$\boldsymbol{w_{i+1}} = \boldsymbol{w_i} + \eta e_i \boldsymbol{u}_i. \tag{3}$$

where $\eta$ indicates the step size.

We now present a comprehensive exposition of the proposed algorithm. The cost function is formulated as the maximum likelihood function of residuals $e_i$, featuring an implicit form

$$J(\boldsymbol{w_i}) = \max\{p(e_i)\}, \tag{4}$$

where $p()$ denotes the PDF. Leveraging the gradient ascent method, the deduced DNN-AF algorithm is expressed in (5)

$$\boldsymbol{w_{i+1}} = \boldsymbol{w_i} - \eta \boldsymbol{u_i} p'(e_i). \tag{5}$$

The nonlinear mapping relationship between $e_i$ and $p'(e_i)$ can be obtained via DNNs, provided the corresponding data is acquired. Given that as $i \to \infty$, $p'(e_i) \to p'(v_i)$, we may effectively employ an approximate representation of the correlation between $v_i$ and $p'(v_i)$ as a feasible alternative [19].

When the noise is wide-sense stationary, multiple viable approaches exist for acquiring background noise information, such as setting the input to zero to directly acquire noise samples [9], [10], [19] and using adaptive cancelers to indirectly obtain noise measurements [22], [23].

Following the acquisition of noise samples, the KDE method can be utilized to reconstruct the underlying PDF of the noise [24]. By superimposing kernel functions centered at each data point, KDE generates a sequence of smoothed probability density estimates, as demonstrated in (6)

$$\hat{p}(v) = \frac{1}{nh}\sum_{i=1}^{n} K\left(\frac{v-v_i}{h}\right), \tag{6}$$

where $n$ denotes the quantity of data samples, $K()$ corresponds to a specific kernel function, and kernel width $h$ determines the smoothness of the density estimation. In this work, we adopt a Gaussian kernel, defined as (7)

$$K(v) = \frac{1}{\sqrt{2\pi}} exp\left\{-\frac{v^2}{2}\right\}. \tag{7}$$

Then by taking the derivative of (6), we obtain the derivative of the underlying PDF .

$$\hat{p}'(v) = \frac{1}{nh^2}\sum_{i=1}^{n} K'\left(\frac{v-v_i}{h}\right). \tag{8}$$

Successful implementation of the above procedures results in two datasets with essential correlation:

1) **Dataset 1**: Measured samples of noise from the system.
2) **Dataset 2**: A corresponding set of PDF derivative values.

Subsequently, a DNN is adopted to establish a sophisticated correspondence relation between the two datasets, which constitutes the core component of the proposed AF framework.

Fig. 1 illustrates the proposed DNN-Driven AF framework.

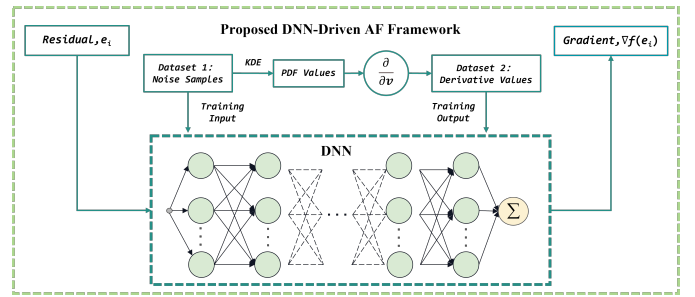

Fig. 1: Proposed DNN-Driven AF Framework

Accounting for the potential non-convexity of the implicit cost function, the filter module requires mandatory pre-training to ensure stable algorithm operation prior to deployment using the industry-standard LMS algorithm [9], [19].

Following the preceding analysis, the DNN-AF algorithm can be formally summarized as **Algorithm 1**.

---

**Algorithm 1** DNN-AF

**1. DNN Preparation**
**Training Input**: Collected noise samples $v_i$ (**Dataset 1**).
**Traing Output**: Corresponding derivative values of the PDF $p'(v_i)$ obtained through the KDE method (**Dataset 2**).
**Objective**: Preparation of a DNN capable of establishing the nonlinear mapping between **Dataset 1** and **Dataset 2** to enable direct gradient acquisition for adaptive filtering.
**2. Initialization**
$w_0$ = a random vector, step size $\eta$, length of pre-training $l$.
**3. Adaptive Filtering**
**For** each instant $i$ **do**
Get the system input $u_i$ and the desired output $d_i$
Compute the estimated error $e_i$ as (2)
   **If** $i < l$ **then**
     Run the LMS algorithm as (3)     **(Pre-Training)**
   **Else**
     Computes $p'(e_i)$ through DNN forward propagation
     Run the proposed algorithm as (5)     **(DNN-AF)**
**End For**

---

## III. PERFORMANCE ANALYSIS

We now develop a convergence analysis for the DNN-AF, examining both mean value and mean square properties. The derivations proceed from the following assumptions.

**Assumption 1:** All the input vectors are temporally independent of each other [25]. That is
$$E\{u_i u_i^T\} = \sigma_u^2 I, \forall i \in \mathbb{N}$$
$$E\{u_i^T u_i\} = M\sigma_u^2, \forall i \in \mathbb{N}$$

**Assumption 2:** The input vector exhibits statistical independence from the noise process at every instant $i$ [25].
$$E\{u_i v_i\} = 0.$$

### A. Mean Value Performance Analysis

The estimated error at instant $i$ is defined by (9)
$$\tilde{w}_i = w_o - w_i. \tag{9}$$

Upon substitution in Eq. (5), we arrive at (10)
$$\tilde{w}_{i+1} = \tilde{w}_i + \eta u_i \left(\frac{p'(e_i)}{e_i}\right) e_i. \tag{10}$$

Proceeding with substitutions into (1) and (2) yields
$$\tilde{w}_{i+1} = \left[I + \eta u_i u_i^T \left(\frac{p'(e_i)}{e_i}\right)\right]\tilde{w}_i + \eta u_i \left(\frac{p'(e_i)}{e_i}\right) v_i.$$

Given that $u_i$ and $v_i$ are both zero mean, so taking the expectation on each side, we have (11)
$$E(\tilde{w}_{i+1}) = E\left[I + \eta u_i u_i^T \left(\frac{p'(e_i)}{e_i}\right)\right] E(\tilde{w}_i). \tag{11}$$

As $i \to \infty$, in the limit where the steady-state error $e_i$ converges to the noise $v_i$, one can obtain
$$\lim_{i\to\infty} E(\tilde{w}_{i+1}) = E\left[I + \eta u_i u_i^T \left(\frac{p'(v_i)}{v_i}\right)\right] \lim_{i\to\infty} E(\tilde{w}_i).$$

Provided that the following condition is satisfied
$$-1 < E\left[1 + \eta \sigma_u^2 \left(\frac{p'(v_i)}{v_i}\right)\right] < 1.$$

Then taking into account **Assumption 1**, we can find that $\lim_{i\to\infty} E(\tilde{w}_i) = 0$ when the step size adheres to (12)
$$0 < \eta < \frac{2}{\sigma_u^2 E\left(-\frac{p'(v_i)}{v_i}\right)}. \tag{12}$$

### B. Mean Square Performance Analysis

Through the application of **Assumption 2**, it becomes mathematically evident that
$$E\left\{\left[I + \eta u_i u_i^T \left(\frac{p'(e_i)}{e_i}\right)\right] \tilde{w}_i \eta u_i \left(\frac{p'(e_i)}{e_i}\right) v_i\right\}$$
$$= \eta E\left[I + \eta u_i u_i^T \left(\frac{p'(e_i)}{e_i}\right)\right] E(\tilde{w}_i) E\left[\frac{p'(e_i)}{e_i}\right] E\{u_i v_i\} = 0.$$

This implies that the cross terms in the mean square expectation vanish mathematically, leading to the observation
$$E\|\tilde{w}_{i+1}\|_2^2 = E\left\{\left[I + \eta u_i u_i^T \left(\frac{p'(e_i)}{e_i}\right)\right]^2\right\}$$
$$\times E\|\tilde{w}_i\|_2^2 + \eta^2 E\{u_i^T u_i\} E\left\{\left[\left(\frac{p'(e_i)}{e_i}\right) v_i\right]^2\right\}. \tag{13}$$

Taking the limit on both sides of (13), one has
$$\lim_{i\to\infty} E\|\tilde{w}_{i+1}\|_2^2 = E\left\{\left[I + \eta u_i u_i^T \left(\frac{p'(v_i)}{v_i}\right)\right]^2\right\} \lim_{i\to\infty} E\|\tilde{w}_i\|_2^2$$
$$+ \eta^2 E\{u_i^T u_i\} E\left\{[p'(v_i)]^2\right\}.$$

Note that $\lim_{i\to\infty} E\|\tilde{w}_i\|_2^2 \to \lim_{i\to\infty} E\|\tilde{w}_{i+1}\|_2^2$, we finally obtain the closed-form expression for the theoretical mean square performance result, shown as (14).
$$\lim_{i\to\infty} E\|\tilde{w}_i\|_2^2 = \frac{\eta^2 M \sigma_u^2 E\left\{[p'(v_i)]^2\right\}}{1 - E\left\{\left[1 + \eta \sigma_u^2 \left(\frac{p'(v_i)}{v_i}\right)\right]^2\right\}}. \tag{14}$$

## IV. NUMERICAL RESULTS

### A. Experimental Parameter Configuration

The results are averaged over 100 independent simulations per iteration. The inputs $u_i$ are 5×1 vectors which follow zero-mean Gaussian distributions. The performance of the adaptive solution is gauged using the mean square deviation (MSD)
$$\text{MSD} = E\left\{\|\tilde{w}_i\|_2^2\right\}. \tag{15}$$

In all scenarios, a dataset of 5,000 pre-collected noise samples served as model inputs. The corresponding PDF derivatives are then computed via the KDE method.

### B. DNN Architectural Configuration

We choose the Multilayer Perceptron (MLP) as a lightweight DNN, achieving substantial computational savings without compromising accuracy. The MLP network architecture used consists of four fully connected layers organized in a progressively contracting pyramidal structure, with each successive hidden layer undergoing exponential dimensionality reduction (from an initial hidden size = 32 down to hidden size = 4). Such architecture is designed to capture complex feature representations while maintaining computational efficienc.

The stochastic gradient descent (SGD) optimizer is used for training the model, with a learning rate of 0.001. The training is performed for 100 epochs with a batch size of 50, with MSE as the loss function.

***Remark 1***. It should be noted that the DNN architecture is not limited to a singular paradigm. The optimal choice of the DNN balancing performance and complexity remains an open problem, which falls beyond the scope of the present work.

### C. Testing Noise Environments

Table II enumerates the noise distributions across diverse environments, where $N()$, $U()$, and $Ray()$ denote Gaussian, Uniform, and Rayleigh distributions, respectively.

TABLE II: Testing Noise Environments

| Type | Distribution |
| --- | --- |
| Impulse | $0.9N(0, 0.1^2) + 0.1N(0, 5^2)$ |
| Uniform | $U(-2, 2)$ |
| Skewed | $Ray(8^2)$ |
| Multi-peak | $0.5N(-3, 2^2) + 0.5N(3, 2^2)$ |

### D. Derivative Fitting Results of Noise PDFs via the DNN

This part presents the DNN's fitting performance on derivatives of noise PDFs, confirming its stable and accurate nonlinear characterization abilities.

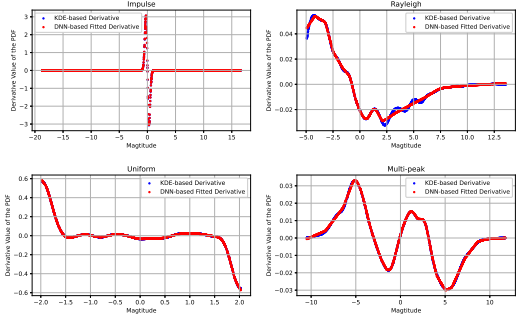

Fig. 2: Derivative Fitting Results of Noise PDFs via the DNN

Fig. 2 presents a comparative analysis of PDF derivative approximation performance across four distinct noise types, where blue points denote KDE-computed derivatives and red points indicate DNN-generated derivative values. The MLP demonstrates excellent fitting accuracy in estimating derivatives for all considered noise variants.

### E. Verification of Theoretical Analysis Results

In this sub-section, we validate the theoretical predictions of the proposed algorithm under environment 1. The TH in the figure represents the mean square value given in Eq. (14).

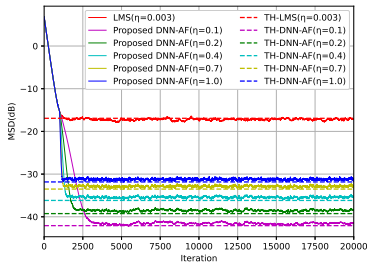

Fig. 3: Transient MSDs (dB) and theoretical steady state MSDs (dB) with different step sizes

As illustrated in Fig. 3, the algorithm demonstrates reliable convergence across all tested step sizes, confirming its mean stability. Furthermore, the close alignment between the steady-state MSD and theoretical values provides compelling evidence that substantiates the mean square performance analysis.

### F. Performance Comparison with Advanced AF Algorithms

In this sub-section, we compare the DNN-AF algorithm with the advanced MCC, MEE, and RBF-AF algorithms in a variety of scenarios. Additionally, in uniform scenarios, the least mean fourth (LMF) algorithm is incorporated into the comparison.

In the impulse scenario, shown as Fig. 4a, narrow kernel widths enable ITL-based methods to demonstrate strong outlier robustness. While the RBF-AF outperforms the ITL-based methods with expanded kernel widths, the DNN-AF algorithm surpasses all competitors, showing superior impact resistance.

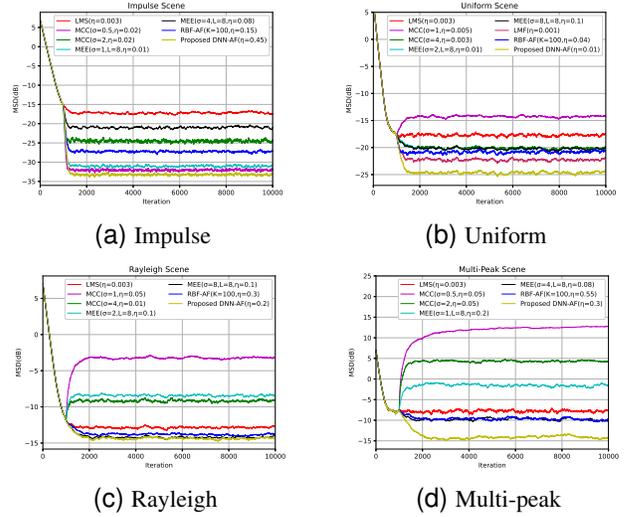

Fig. 4: Performance comparison under various scenes

In the Uniform scenario, both the MCC and MEE algorithms exhibit performance degradation. Remarkably, the DNN-AF algorithm demonstrates the most favorable steady-state MSD performance, surpassing even the LMF algorithm.

In the Rayleigh scenario, affected by skewed characteristics, existing algorithms degrade significantly. While the RBF-AF algorithm shows marginal improvement over the LMS, the proposed DNN-AF algorithm demonstrates superior performance, as presented in Fig. 4c.

In the Multi-peak scenario, ITL-based algorithms deteriorate further. The RBF-AF algorithm degenerates to the level comparable with the MEE, while the DNN-AF algorithm maintains exceptional robustness, demonstrating unequivocal superiority over all competing methods, as depicted in Fig. 4d.

Figs. 4a-4d collectively demonstrate that competing algorithms struggle with performance consistency under varying noise conditions. While the RBF-AF offers modest generalization, the proposed DNN-AF excels across all scenarios, providing compelling evidence that DNN-based AF architectures can deliver superior generalization capability.

## V. CONCLUSION

This letter proposes a DNN-driven framework to address the longstanding generalization challenge in AF. DNNs are structurally embedded in the core architecture of the AF system. Parallel to the classical AF framework's focus on the design of the cost function, the proposed framework specializes in direct gradient acquisition, achieving a genuinely model-free architecture. Numerical results demonstrate that the DNN-AF algorithm delivers enhanced generalization capacity in various non-Gaussian scenes, surpassing state-of-the-art methods.

The proposed framework has many expansion possibilities. By varying training objectives and DNN architectures, a family of AF algorithms may emerge, with natural extensions to related domains in signal processing and learning systems. This will be a key focus for our future investigation.